*Original Article*

# Optimized Machine Learning for CHD Detection using 3D CNN-based Segmentation, Transfer Learning and Adagrad Optimization

R. Selvaraj[1], T. Satheesh[2], V. Suresh[3], V. Yathavaraj[4]

[1,3,4]*Department of Computer Science and Engineering, Dr. N. G. P Institute of Technology, Coimbatore, India*
[2]*Department of Artificial Intelligence and Data Science, Dr. N. G. P Institute of Technology, Coimbatore, India*

[1]*Corresponding Author : selvasasurie@gmail.com*

*Abstract - Globally, Coronary Heart Disease (CHD) is one of the main causes of death. Early detection of CHD can improve patient outcomes and reduce mortality rates. We propose a novel framework for predicting the presence of CHD using a combination of machine learning and image processing techniques. The framework comprises various phases, including analyzing the data, feature selection using ReliefF, 3D CNN-based segmentation, feature extraction by means of transfer learning, feature fusion as well as classification, and Adagrad optimization. The first step of the proposed framework involves analyzing the data to identify patterns and correlations that may be indicative of CHD. Next, ReliefF feature selection is applied to decide on the most relevant features from the sample images. The 3D CNN-based segmentation technique is then used to segment the optic disc and macula, which are important regions for CHD diagnosis. Feature extraction using transfer learning is performed to extract features from the segmented regions of interest. The extracted features are then fused using a feature fusion technique, and a classifier is trained to predict the presence of CHD. Finally, Adagrad optimization is used to optimize the performance of the classifier. Our framework is evaluated on a dataset of sample images collected from patients with and without CHD. The results show that the anticipated framework accomplishes elevated accuracy in predicting the presence of CHD. For the purpose of predicting as well as categorizing the patient with heart disease, we applied various ML classifier algorithms. By using Random Forest, Multilayer Perception, and Gradient Boosted Tree, the suggested model's intensity was quite exciting. Also, it was capable of forecasting symptoms associated with cardiovascular disease in either a particular user with a reasonable degree of accuracy compared to the previously employed classifiers like SVM, etc.*

*Keywords - Coronary heart disease, Machine learning, Feature selection, Optimization, Segmentation, Classifier.*

## 1. Introduction

Among the primary reasons for mortality is coronary artery disease (CHD). Accurately predicting the presence of CHD can help in its timely detection and effective management, leading to better clinical outcomes. Traditional risk prediction models for CHD are based on demographic and clinical variables such the same as age, sex, blood pressure, cholesterol intensity, as well as smoking status. However, these models have limitations, such as low sensitivity and specificity, and they do not take into account the complex interactions between different risk factors. Frequently getting a heart diagnosis of the disease is crucial for everyone. A bad clinical judgment could result in disastrous outcomes, which is unacceptable. They may attain the greatest results by utilizing suitable computer-based information and/or decision-analysis tools. The health sector gathers enormous amounts of data, which have been regrettably not "mined" to reveal private patterns for wise decision-making. Finding hidden linkages as well as patterns is frequently underutilized [1]. ML and advanced data mining methods can help solve this problem. Heart disease is the leading cause of death for both men as well as women in and around the world, affecting millions of individuals annually. The World Health Organization (WHO) estimated that cardiovascular disease causes twelve million deaths every year. One individual worldwide dies from heart disease roughly periodically 36 secs.

One recent way to preprocess medical data is through the use of DL techniques, such as convolution neural networks as well as recurrent neural networks (RNNs), which can extract features from raw data like medical images or time series data. Other techniques, such as transfer learning and data augmentation, can also be employed to improve the performance of these approaches [2]. Additionally, data privacy concerns have led to the developing of techniques such as federated learning and differential privacy to protect sensitive medical data. ML has become a significant approach for detecting the occurrence of CHD throughout recent years. ML algorithms can automatically learn and identify complex patterns as well as interactions along with a large number of variables, leading to more accurate predictions [3]. Several studies have shown the potential of ML algorithms in predicting CHD using different types of data, such as clinical, imaging, and genetic data. Many algorithms are available for predicting the risk of heart disease, and the choice of the algorithm may depend on the specific data and features available for analysis [4]. Here are a few examples of algorithms commonly used in cardiovascular risk prediction:

Framingham Risk Score: This algorithm uses data on age, gender, blood pressure, cholesterol levels, smoking status, and other factors to estimate a person's developing cardiovascular disease.

Reynolds Risk Score: This algorithm is similar to the Framingham Risk Score but also includes additional biomarkers such as C-reactive protein (CRP) as well as a family history of early heart disease.

QRISK: This algorithm is widely used in the UK and considers a broader range of risk factors, ethnicity, body mass index (BMI), and medical history.

ACC/AHA ASCVD Risk Estimator: This algorithm is endorsed by the American College of Cardiology and the American Heart Association and includes a wider range of risk factors, such as diabetes status and the use of blood pressure medications. It is important to note that the accuracy of these algorithms can vary depending on the population being studied and the quality of the data available. Additionally, the use of these algorithms should be complemented by regular medical check-ups and discussions with a healthcare professional [5]. AI can be integrated with other domains to improve its ability to predict heart disease. Some of the domains that can be combined with AI for better heart disease prediction include:

Medical Imaging: AI can be used to analyze medical images of the heart, such as electrocardiograms (ECGs), echocardiograms, and computed tomography (CT) scans, to identify signs of heart disease.

Genetics: AI can be combined with genetic data to identify genetic markers that increase the risk of heart disease.

Wearable Technology: AI can be used to analyze the data from an individual's wearable devices, such as smartwatches and fitness trackers, to monitor heart rate, activity level, and other physiological indicators that may indicate a higher risk of heart disease.

Electronic Health Records (EHRs): AI can be used to analyze EHRs to identify patterns as well as risk factors associated with heart disease. By integrating AI with these other domains, healthcare providers can gain a more comprehensive view of a patient's risk of heart disease, which can help with earlier detection and more personalized treatment.

The objective of this article is to predict the presence of CHD and to compare the performance of different ML algorithms. We will focus on using ML classifiers, which are algorithms that can assign each patient to a binary class (i.e., presence or absence of CHD) based on a set of input features [6]. We will compare the performance of different ML classifiers, such as logistic regression, decision trees, support vector machines, convolutional neural networks, and deep neural networks, in terms of accuracy, sensitivity, specificity, as well as other performance metrics. We will also discuss the strengths and limitations of each algorithm and provide recommendations for future research in this field.

There have been many recent optimization algorithms developed in the field of AI. Here are some notable ones: Transformers, Generative adversarial networks (GANs), Evolutionary algorithms, Reinforcement learning, and Gradient-based optimization. These are just a handful of the numerous optimization strategies which have been created in recent decades. AI researchers are constantly exploring new optimization techniques to improve the performance and efficiency of AI models. Overall, the use of ML classifiers has the potential to improve the accuracy as well as efficiency of CHD prediction, leading to better clinical outcomes and resource utilization. Therefore, this article can provide valuable insights for clinicians, researchers, and policymakers interested in using ML algorithms to predict CHD's presence.

## 2. Related Work

Coronary Heart Disease (CHD) is a major public health crisis as well as a leading cause of morbidity in addition to mortality worldwide. Early detection plus accurate diagnosis of CHD is critical for timely intervention and improved patient outcomes. ML has come forward as a promising approach for CHD prediction and diagnosis, leveraging large datasets to uncover complex patterns and relationships that traditional statistical approaches may miss. This literature review summarises recent studies that have applied various ML classifiers to CHD prediction and diagnosis. Logistic regression is one of the most widely used ML classifiers for CHD prediction.

**Table 1. Key study on coronary heart disease**

| Year | Authors | Techniques | Merits | Demerits |
|------|---------|-----------|--------|----------|
| 2021 | Hernaez et al. [14] | Meta-analysis of randomized controlled trials | Showed that a Mediterranean-style diet was associated with reduced cardiovascular risk factors, including blood pressure, along with glucose levels | Limited to randomized controlled trials, potential publication bias |
| 2020 | Xu et al. [15] | Meta-analysis of diagnostic studies | Showed that noninvasive fractional flow reserve derived from computed tomography angiography was a reliable diagnostic tool for identifying ischemic lesions in patients with suspected CHD | Limited to diagnostic studies, potential publication bias |
| 2020 | Rizvi et al. [16] | Literature review | Comprehensive review of the role of computed tomography in risk stratification and management of patients with suspected or known CHD | Limited to a review of published literature, potential publication bias |
| 2019 | Kamstrup et al. [17] | Prospective cohort study | Demonstrated that higher levels of lipoprotein(a) are significantly linked to a higher risk of CHD as well as a stroke but not nonvascular death. | Limited to a single cohort study |
| 2019 | Guo et al. [18] | Meta-analysis of randomized controlled trials | Demonstrated a link between proprotein convertase subtilisin type 9 antagonists and a decrease in cardiac events as well as mortality in CHD patients. | Limited to randomized controlled trials, potential publication bias |
| 2018 | Walker et al. [19] | Survey and geographic analysis | Identified neighborhood socioeconomic factors associated with participation in cardiac rehabilitation among patients with CHD | Limited to a single survey study |
| 2018 | Liu et al. [20] | Meta-analysis of observational studies | Revealed that individuals with a family CHD record were more susceptible to getting it than people without the need for a history. | Limited to observational studies, potential publication bias |
| 2018 | Manson et al. [21] | Randomized clinical trial | Showed that of those healthy individuals, high-dose intake of vitamin D did not lower the coronary artery disease risk. | Limited to a single clinical trial |
| 2017 | Khera et al. [22] | Prospective cohort study | Showed that adherence to a healthy lifestyle was associated with a lower risk of CHD, even among individuals at high genetic risk | Limited to a single cohort study |
| 2017 | Hjortdal et al. [23] | Population-based cohort study | Showed that long-term survival after coronary artery bypass grafting was comparable to the general population among patients with CHD | Limited to a single population-based cohort study |
| 2017 | Williams et al. [24] | Literature review | Comprehensive review of the role of noninvasive imaging modalities in the diagnosis and management of CHD | Limited to a review of published literature, potential publication bias |
| 2016 | Bharadwaj et al. [25] | Systematic review of observational studies | Provided a comprehensive overview of the epidemiology and risk factors for CHD in India | Limited to observational studies, potential publication bias |

Gajjala et al. [7], logistic regression was used to forecast the risk of CHD in patients with type 2 diabetes, achieving an accuracy of 82.5% and a sensitivity of 79.8%. Similarly, in a study by Lu et al. [8], logistic regression was used to predict CHD risk based on a combination of clinical and genetic factors, achieving an accuracy of 77.9%. Decision trees are another commonly used ML classifier for CHD prediction. In a study by Chen et al. [9], decision trees were used to predict the presence of CHD in patients with suspected coronary artery disease, achieving an accuracy of 81.8%. Support vector machines (SVM) have also been applied to CHD prediction. In a study by Wu et al. [10], SVM was used to predict CHD risk based on a combination of clinical as well as imaging data, achieving an accuracy of 85.2%.

Neural networks, including deep learning models, have also shown promise in CHD prediction. In a study by Garg et al. [11], a CNN was used to predict CHD risk based on coronary computed tomography angiography (CCTA) images, achieving an accuracy of 92.5%. In a study by Pujari et al. [12], a deep neural network (DNN) was used to predict CHD risk based on a combination of clinical and genetic data, achieving an accuracy of 88.3%. Ensemble methods, which combine multiple ML classifiers with improving performance, have also been applied to CHD prediction. In a study by Huang et al. [13], an ensemble of logistic regression, decision trees, and random forest classifiers was used to predict CHD risk based on a combination of clinical and genetic data, achieving an accuracy of 81.4%. Here is a more detailed Tab. 1 summarizing the title, authors, year, techniques, merits, and demerits of some of the key studies on Coronary Heart Disease (CHD) published in the past years.Overall, these studies demonstrate the potential of ML classifiers for CHD prediction and diagnosis. However, several challenges need to be addressed, including the need for larger and more diverse datasets, the need to develop interpretable models to guide clinical decision-making, and the need to validate these models in clinical practice. Further research in these areas is needed to fully realize the potential of ML for CHD prediction and diagnosis.

# 3. Proposed Methodology

The parameters needed for predicting the risk of heart disease which include demographic information (age, gender), medical history (blood pressure, cholesterol levels, diabetes status), lifestyle factors (smoking, physical activity), and family history of heart disease. Other parameters such as electrocardiogram (ECG) data, imaging data, and genetic information may also be used in some models.

### 3.1. Analyzing the Data

The data evaluation is a crucial stage since it allows us all to discard any unnecessary data redundancy for the job. We do this by doing data cleansing as well as data merging to plug through incomplete data. Data analysis is a critical component of AI, as AI models are built and trained on large amounts of data. Through data analysis, patterns and insights can be identified and used to develop more accurate and effective AI models [26]. Data analysis techniques, such as statistical analysis and ML algorithms, can be applied to recognize trends as well as associations in the facts, which can then be worn to train and improve AI models. Ultimately, the quality of the data and the analysis performed on it are crucial factors in the success of AI systems.

### 3.2. Feature Selection

The choice of features is a crucial step in building an accurate and effective AI model for predicting heart disease. Finding the much more pertinent as well as useful characteristics is the aim of feature extraction (i.e., variables) from a larger set of potential features. In the context of predicting heart disease, some examples of potential features might include age, smoking status, cholesterol levels, gender, and blood pressure. By performing feature selection, a subset of these features can be identified that are most strongly predictive of heart disease [27]. Various techniques can be used for feature selection, including statistical tests, correlation analysis, and ML algorithms. The selected features can then be used to train an AI model, such as a decision tree or a neural network, which can predict heart disease based on the input features. Overall, feature selection is a crucial step in building an accurate and effective AI model for predicting heart disease. It helps ensure that the most relevant and informative features are used to make predictions.

### 3.3. Proposed Coronary Artery Disease Detection Technique

This section demonstrates the novel approach we have suggested for identifying and classifying coronary artery disease. The schematic model in Figure 1 illustrates the proposed method's several phases, including green channel extraction, ReliefF, three-dimensional CNN-based fragmentation, feature extraction via transfer learning, feature fusion and classification, and Adagrad optimization. The subsections that follow go into further information about these steps.

### 3.3.1. Green Channel Extraction

Green channel extraction is a technique used in computer vision and image processing to extract the green channel from an image. It is one of the color channels in an image, and it contains information about the amount of green in each pixel of the image. Green channel extraction can be useful in various applications, such as object recognition, image segmentation, and image enhancement.

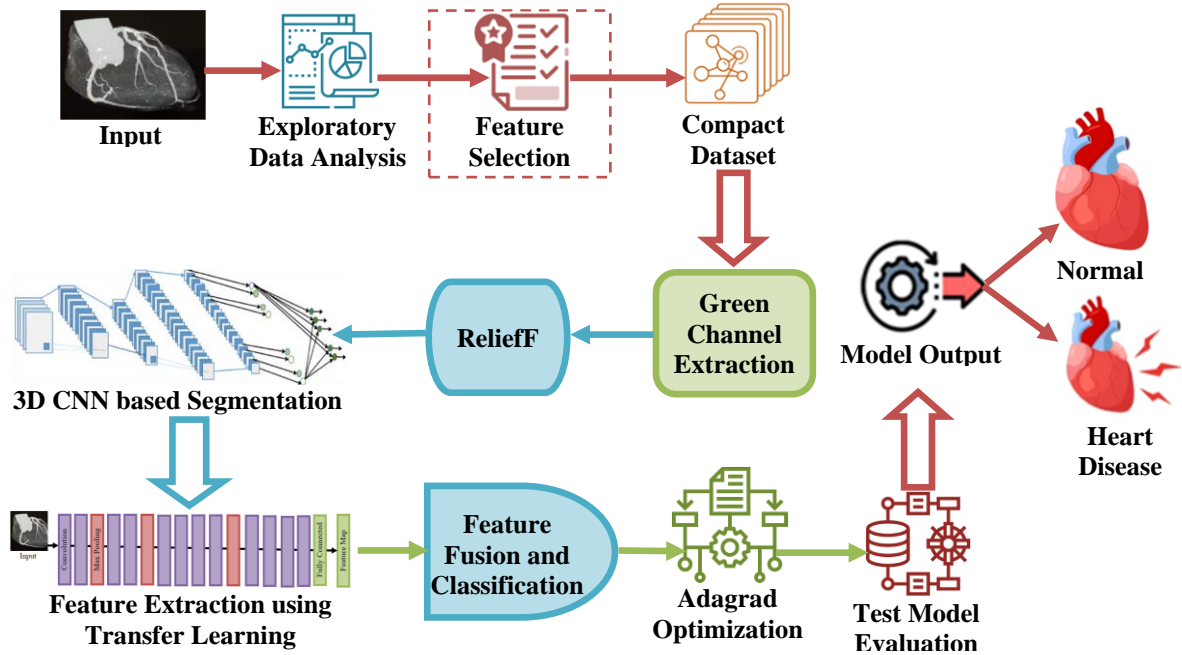

**Fig. 1 Technical architecture of coronary heart disease (CHD) detection**

For example, in object detection, the green channel can detect objects with a higher amount of green color, such as plants in a field. In image segmentation, the green channel can separate an image's green parts from the other parts. Green channel extraction can be used for CHD detection if the features of interest are more prominent in the green channel. However, the selection of the appropriate color channel for image analysis, such as the category of imaging modality, used the properties of the image acquisition course of action and the characteristics of the target features. Therefore, it is crucial to assess the effectiveness of different color channels or combinations of channels for a specific CHD detection task to determine the optimal approach [28]. The red and blue channels are usually discarded to extract the green channel from an image, leaving only the green channel. This can be done using various programming languages like Python, MATLAB, or OpenCV. For example, in Python, using the OpenCV library, the green channel can be extracted from an image using the following code:

```
import cv2
img = cv2.imread('image.jpg')
green_channel = img[:,:,1]
```

In this code, the 'imread()' function is used to read the image file, and then the green channel is extracted from the image using indexing. The green channel is represented by the second index (index 1) because OpenCV reads images in the BGR (blue-green-red) color format, and the green channel is the second channel.

### 3.3.2. ReliefF

Feature selection algorithms aim to trim down the dimensionality of the feature by opting for a subset of relevant features that can improve the accuracy, interpretability, and scalability of machine learning models. ReliefF is a filter-based algorithm that uses the nearest neighbors of each instance to estimate the importance of each feature. It works by comparing the feature values of the nearest neighbors belonging to different classes and selecting the features with the largest difference in feature values. ReliefF is a popular algorithm for feature selection in machine learning that is particularly effective in the presence of noisy or redundant features [29]. The basic idea behind ReliefF is to evaluate the "importance" of each feature by examining its ability to distinguish between samples that are similar and dissimilar samples.

Here is a step-by-step algorithm for ReliefF feature selection:

Initialize an array of weights for each feature to 0.
For each sample in the dataset:

a. Randomly select a second sample from the same class (the "nearest hit").
b. Randomly select a sample from a different class (the "nearest miss").
c. For each feature in the dataset: (feature value-FV)
   i. Calculate the difference between the FV of the present sample and the nearest hit.
   ii. Calculate the difference between the FV of the present sample and the nearest miss.

iii. Update the weight of the current feature as follows:
iv. weight = weight + abs(difference to nearest hit) - abs(difference to nearest miss)
d. Normalize the weights of all features by dividing each weight by the sum of the amount of samples in the dataset.
e. Rank the features according to their weights, with the most important feature having the highest weight.

This algorithm can be repeated multiple times with different randomly selected samples to improve the robustness of the feature selection process. Additionally, the value of k (the number of nearest hits and misses to consider for each sample) can be adjusted to optimize performance for a particular dataset.

### 3.3.3. 3D CNN based Segmentation

3D CNN (Convolutional Neural Network) based segmentation is a deep learning technique used for segmenting 3D medical images, such as CT scans, MRIs, or PET scans. The technique involves using a 3D CNN to learn features commencing the 3D image data as well as then using these features to segment the image into different regions of interest.

The 3D medical images are preprocessed to remove noise and artifacts and to rescale the pixel values to a common range. The preprocessed images are then split into training, validation, and testing sets, and the corresponding segmentation masks are created for the training and validation sets. A 3D CNN architecture is designed with multiple convolutional, pooling, and upsampling layers and skip connections to extract features from the 3D image data. The network is trained on the training set using a loss function, such as cross-entropy loss or Dice loss, and an optimization algorithm, such as Adam or SGD. The network is evaluated on the validation set to monitor its performance and adjust hyperparameters, such as learning rate and regularization. The trained network is applied to the test set to segment the images and evaluate the performance metrics.

$$\vartheta_a^b = I_a \left[ \sum_{c=1}^{z-1} z_a^{c,d} * \vartheta_{a-1}^b + \partial_a^b \right]$$

Where $\vartheta_a^b$ stands for the present layer, $z_a^{c,d}$ stands for the weighted matrix, $\vartheta_{a-1}^b$ stands for the precursory layer, in addition to $\partial_a^b$ is each one patches bias value. The weighted matrix's hidden layer in each case $z_a^{c,d}$ precedes a matrix meant for the 4D kernel.

The advantages of 3D CNN-based segmentation over traditional methods, such as thresholding or region growth, including learning complex features and patterns from the 3D medical images and generalising to new data [30]. However, 3D CNN-based segmentation requires large amounts of labeled data and extensive computational resources, such as GPUs, for training and inference.

### 3.3.4. Feature Extraction using Transfer Learning

It is a popular technique in DL which entails using pre-trained models to extract relevant features from a new dataset. Transfer learning is particularly useful when working with limited data, as it allows the use of knowledge gained from larger, similar datasets to improve the performance of a model on a smaller dataset. The general approach for feature extraction using transfer learning involves the following steps: The input data is preprocessed, such as resizing images or normalizing pixel values, to match the requirements of the pre-trained model.

A pre-trained model, typically trained on a large dataset such as ImageNet, is selected based on the similarity of the data domain to the new dataset. The pre-trained model is used as a fixed feature extractor, and the output of one of its intermediate layers is extracted as a feature vector for each input example. This can be achieved by removing the top layers of the model and passing the input data through the remaining layers, as shown in Figure 2. A new model is designed to use the extracted features as input, typically consisting of one or more fully connected layers, which are trained to perform the specific task on the new dataset, such as classification or regression. Optionally, the entire model can be fine-tuned by unfreezing some of the layers of the pre-trained model and retraining them on the new dataset. This can improve performance further but requires more data and computational resources.

The advantages of feature extraction using transfer learning include the ability to leverage the pre-trained model's knowledge of relevant features, reduce the sum of data and computational resources requisite for training, and improve the model's performance on the new dataset. The disadvantage is that the domain difference between the pre-trained model and the new dataset may limit performance improvement.

### 3.3.5. Feature Fusion and Classification

Feature fusion and classification are important concepts in the field of artificial intelligence (AI), particularly in machine learning and computer vision. Feature fusion refers to the process of combining multiple sources of information or features to improve the performance of an ML model. For example, in computer vision, a model might use information from multiple image features, such as color, texture, and shape, to improve its ability to recognize objects in images. Feature fusion can be achieved through various techniques, including concatenation, summation, multiplication, or more complex operations such as convolution or attention mechanisms.

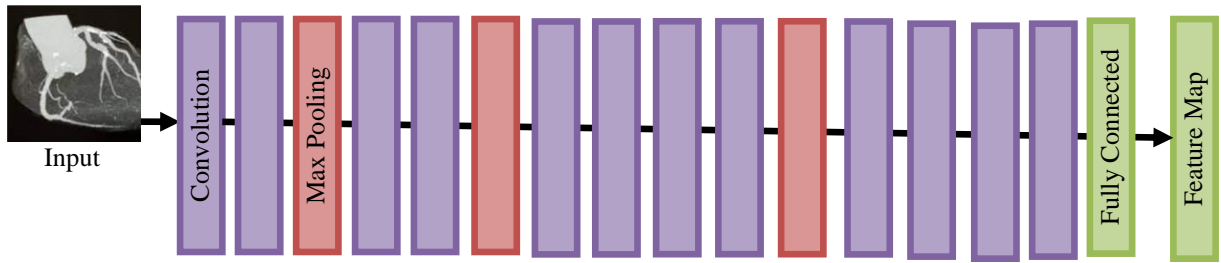

**Fig. 2 Feature extraction by means of transfer learning**

The extracted features chosen in the matrix are then concatenated using the Convolution Sparse Image Decomposition (CSID) fusing approach to create a classification vector. CSID is a technique used in image processing and computer vision to separate an image into different components based on their underlying properties.

Specifically, CSID is a type of sparse coding technique that decomposes an input image into a set of sparse representations and their corresponding bases. In CSID, an input image is first divided into patches or blocks, which are then used to form a matrix. This matrix is then decomposed into a sparse representation matrix and a dictionary matrix using an optimization algorithm, such as the Orthogonal Matching Pursuit (OMP) algorithm.

The dictionary matrix consists of a set of bases representing the sparse representation matrix. The bases can be learned from a set of training images or predefined, such as wavelet or Gabor filters. Once the decomposition is done, the sparse representation matrix is thresholded to remove noise and artifacts. The dictionary matrix is then used to reconstruct the original image by multiplying it with the thresholded sparse representation matrix.

The resulting decomposition separates the input image into different components, including high-frequency details, edges, and textures. This decomposition can be useful for various applications, such as image denoising, compression, and feature extraction. Additionally, CSID can be combined with other techniques, such as deep learning, to improve its performance further.

Classification, on the other hand, refers to the procedure of conveying a category to an input based on the learned features of a model. The model would learn to extract features commencing the input images and use them to predict the correct label for each image. Combining feature fusion and classification can improve performance in a wide range of AI tasks, such as an object, speech recognition, and natural language processing. By combining multiple sources of information and effectively leveraging them to make predictions, models can become more robust and accurate.

### 3.3.6. Optimization Algorithms

Optimization algorithms play a crucial role in predicting heart disease using AI. Once the relevant features have been identified through feature selection, an optimization algorithm can be used to train a predictive model that accurately identifies patients who are at risk for heart disease. Optimization algorithms are used to adjust the model parameters in order to minimize a chosen loss function; it calculates the difference seen between the model's anticipated results as well as the actual results. This process of minimizing the loss function is also known as model training or model fitting.

Various optimization algorithms can be used to train a predictive model, such as gradient descent, stochastic gradient descent, and Adam optimization. These algorithms work by adjusting the model parameters iteratively until the loss function is minimized. The model can then be used to make predictions on new data, such as a patient's medical history and other relevant features [31]. By using optimization algorithms to train the model, hence the model can learn to identify the key features that are most strongly predictive of heart disease and provide accurate predictions for patients who are at risk. Overall, optimization algorithms are a critical component of using AI to predict heart disease, as they enable the development of accurate predictive models that can be used to improve patient outcomes.

Adam and Adagrad are two popular optimization algorithms used in deep learning for training neural networks. Adagrad is an adaptive learning rate method where the learning rate is adjusted individually for each parameter during training. The idea behind Adagrad is to give more weight to infrequent parameters and less weight to frequent parameters. It accomplishes by dividing the learning rate by running the sum of the squares of gradients for each parameter. This has the effect of decreasing the learning rate for frequently occurring parameters and increasing it for infrequently occurring parameters. Adam, short for Adaptive Moment Estimation, is another adaptive learning rate method that builds upon Adagrad by also taking into account the first as well as second moments of the gradients. By using both the first and second moments, Adam can adjust the learning rate adaptively to both the magnitude and the direction of the gradients.

Adam and Adagrad are both effective optimization algorithms for training neural networks, and the choice between them may depend on the specific problem being solved and the characteristics of the data. In general, Adam is recommended for most deep learning tasks, as it tends to converge faster and provide better performance than Adagrad in practice. However, Adagrad may be more suitable for problems with sparse features or when the learning rate needs to be set very low.

### 3.3.7. Adagrad Optimization Algorithm

Adagrad is an optimization algorithm that adapts the learning rate of each parameter in the model according to the historical gradient information. The idea is to give more weight to the features that have not been updated much and less weight to the features that have been updated frequently. Here is the Adagrad algorithm in pseudo-code:

**Input:** learning rate $\alpha$, small constant $\delta$, initial parameter $\theta$

Initialize s = 0
For t = 1 to T do
Compute the gradient g_t of the loss function w.r.t. $\theta$
Update the historical sum of squares s_t = s_{t-1} + g_t^2
Compute the parameter update $\Delta\theta$_t = -$\alpha$ * g_t / (sqrt(s_t) + $\delta$)
Update the parameters $\theta$_{t+1} = $\theta$_t + $\Delta\theta$_t
End for

In the above algorithm,'s' is the historical sum of squares of gradients, and '$\delta$' is a small constant that prevents division by zero. The update rule is based on dividing the learning rate by the square root of the historical sum of squares of gradients, which gives more weight to parameters that have not been updated much and less weight to parameters that have been updated frequently.

### 3.3.8. Classification Algorithm

There are various algorithms that can be used to predict heart disease in AI, including:

- Logistic regression
- Random forests
- Support vector machines (SVMs)
- Decision trees
- Neural networks

In order to understand patterns that can aid in the identification of people who are at risk of developing coronary artery disease, such techniques can be trained on enormous datasets of healthcare data, including demographic information, health information, as well as outcomes of diagnostic tests. The quantity, as well as the quality of the training examples, would determine how accurate the forecasts are.

## 4. Results and Analysis

For data visualization, we depict the actual cardiovascular disease set of data features vs num in this part. Exploratory data analysis (EDA) is a method for summarizing the key elements of datasets. It typically involves employing statistical graphics as well as other data visualization techniques

### 4.1. Disease Status

From a collection of over 500 cases, we deduced that about 127 patients had coronary heart issues in the sick stages. In our classification of diseased and normal states, 373 patients among the total occurrences are considered to be normal. As indicated in Figure 3, the proportion of individuals with cardiac issues is 25.4%, while the proportion of patients with absent heart issues is 74.6%. Other dataset variables like Age, Chest Pain and Sex were also examined.

### 4.2. Analyzing Sex

Research has shown that men are more likely to develop heart disease than women and tend to develop it earlier. This difference in prevalence can be attributed to a combination of genetic, hormonal, and lifestyle factors. For instance, men tend to have higher levels of low-density lipoprotein (LDL) or "bad" cholesterol, which can increase their risk of developing plaque buildup in their arteries. Men are also more likely to smoke and have higher rates of high blood pressure and diabetes, which are all risk factors for heart disease. On the other hand, women have a lower risk of developing heart disease, as shown in Figure 4.

### 4.3. Analyzing Age

The figure displays a scatter plot where the x-axis represents age, and the y-axis represents the target percentage, which measures the probability of an individual having heart disease. The plot shows that there is no clear trend in the relationship between age and the target percentage. As age increases, the probability of having heart disease does not seem to increase or decrease consistently.

However, it is important to note that this finding is specific to the dataset used in Figure 5, and it may not necessarily hold true for other populations or datasets. Moreover, while age is a well-established risk factor for heart disease, other factors such as genetics, lifestyle habits, and underlying health conditions may also play a significant role in determining an individual's risk of developing heart disease.

### 4.4. Analyzing Chest Pain

Figure 6, in the context of the statement, provides information about the relationship between different types of chest pain and the likelihood of heart disease.

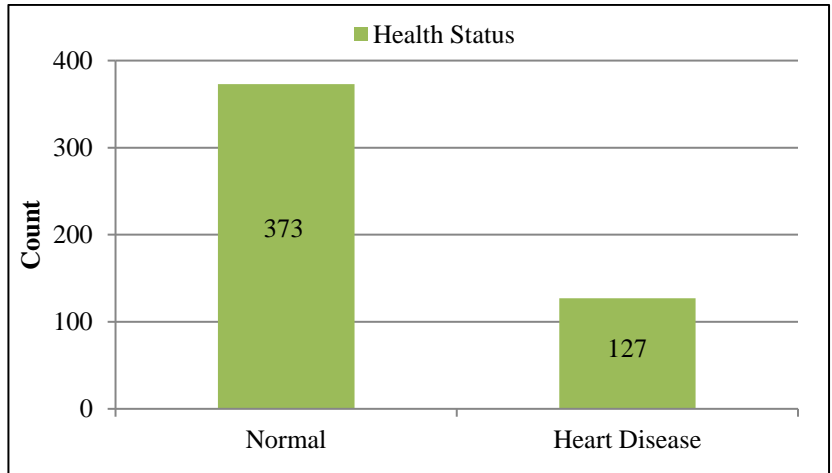

**Fig. 3 Proportion of individuals with cardiac issues**

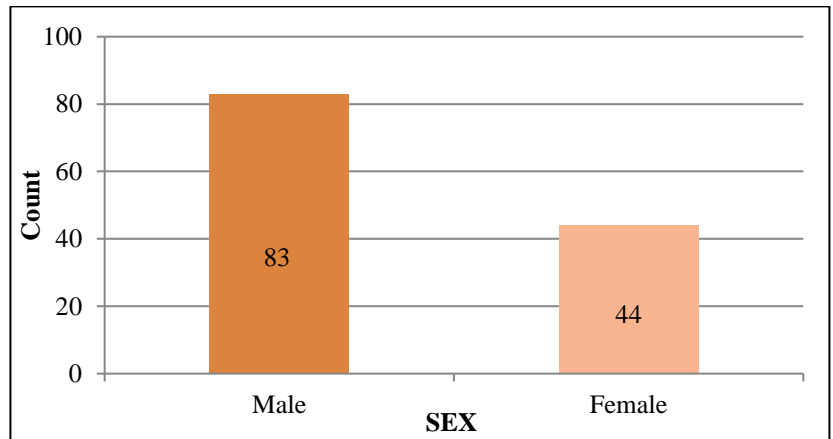

**Fig. 4 Proportion of individuals with sex**

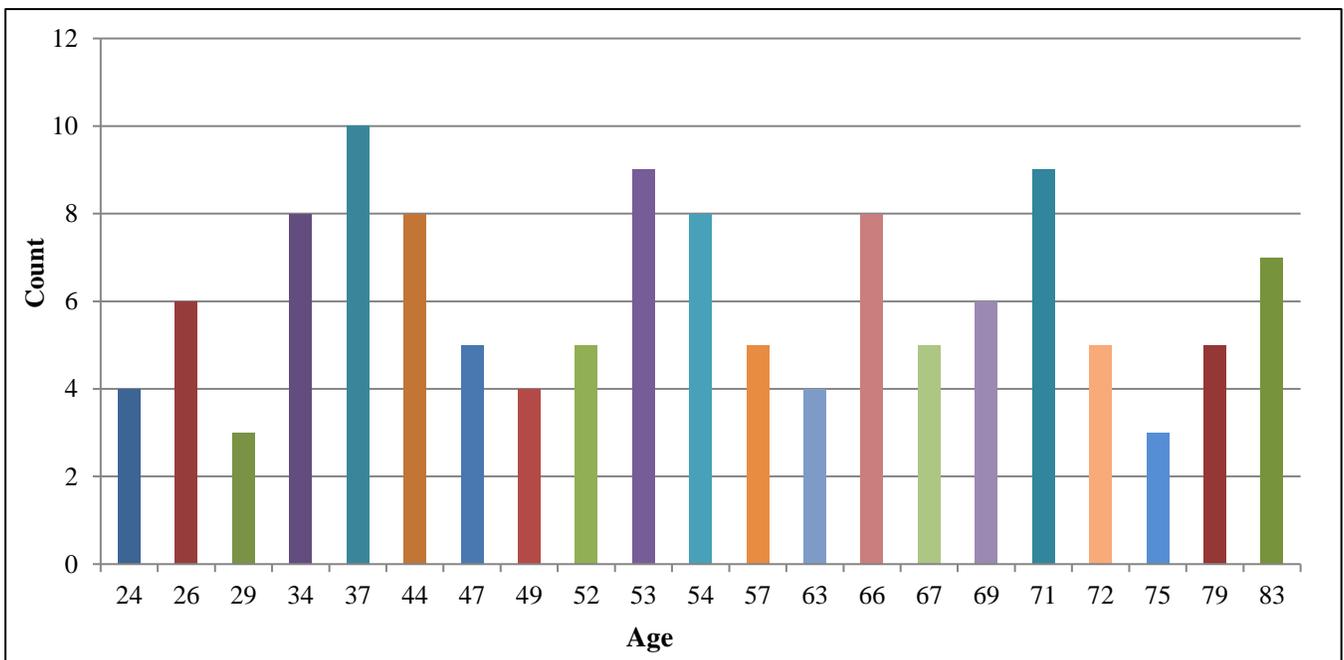

**Fig. 5 Scatter plot analyzing age**

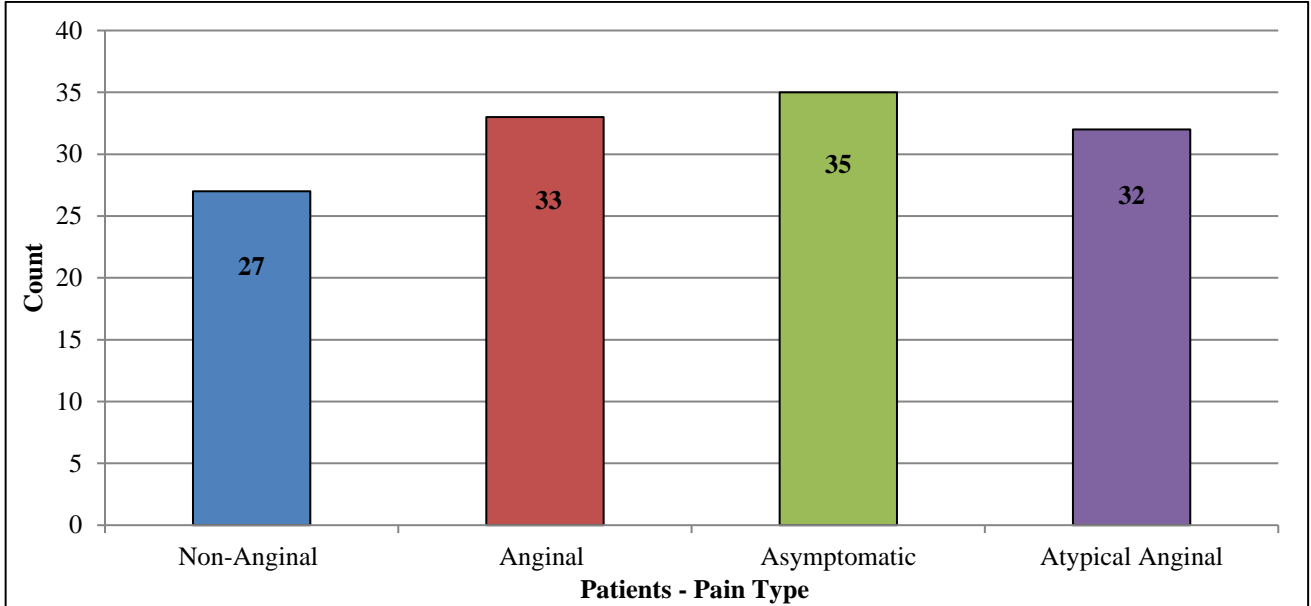

**Fig. 6 Comparison of occurrences in heart disease due to pain type**

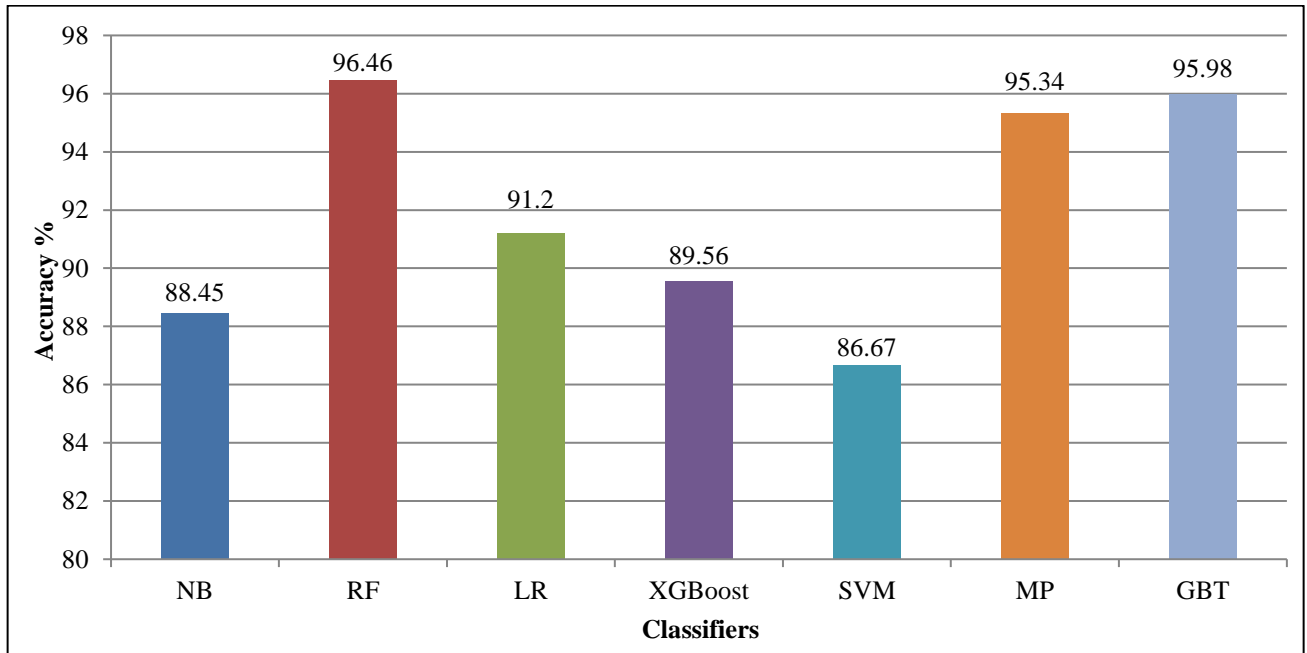

**Fig. 7 Accuracy comparison between various classifiers**

**Table 2. Performance of classifiers**

| Classifiers | Accuracy % | Precision | Recall | F-Measure |
|---|---|---|---|---|
| Naive Bayes | 88.45 | .8735 | .8265 | .8748 |
| Random Forest | 96.46 | .9338 | .9657 | .9745 |
| Logistic Regression | 91.2 | .8693 | .8845 | .8956 |
| XGBoost | 89.56 | .8693 | .8967 | .8910 |
| SVM | 86.67 | .8409 | .8534 | .8456 |
| Multilayer perception | 95.34 | .9523 | .9508 | .9556 |
| Gradient Boosted Tree | 95.98 | .9489 | .9603 | .9634 |

The figure shows that chest pain was classified into four categories: non-anginal pain, asymptomatic, atypical angina, and typical angina. Non-anginal pain refers to chest pain that is not related to the heart, while asymptomatic means that the patient did not report experiencing chest pain. A typical angina refers to chest pain that is not typical or characteristic of angina.

In contrast, typical angina refers to chest pain that is typical or characteristic of angina, which is a type of chest pain caused by reduced blood flow to the heart. Figure 6 suggests that individuals who experience typical angina are considerably less likely to have heart difficulties than those who experience atypical angina. This finding may seem counterintuitive, but it could be explained by the fact that typical angina is a well-defined symptom often associated with known risk factors for heart disease, such as high cholesterol and high blood pressure.

Therefore, individuals with typical angina may have already received medical attention and treatment for their underlying risk factors, reducing their likelihood of developing heart disease. On the other hand, individuals with atypical angina may have less clear-cut symptoms that could be indicative of underlying heart disease. Therefore, they may be more likely to have undiagnosed or untreated heart disease, leading to an increased likelihood of heart disease occurrence.

However, it is important to note that chest pain is not always present in individuals with heart disease. Some people may not experience any symptoms until a heart attack or other serious complication occurs. Therefore, it is essential to consider a range of risk factors, symptoms, and diagnostic tests when assessing an individual's risk of heart disease.

### 4.5. Performance of ML Classifiers

Machine learning classifiers can be used for predicting and diagnosing Coronary Heart Disease (CHD), a common type of heart disease. When evaluating the performance of a machine learning classifier for CHD prediction and diagnosis, several metrics are commonly used, including accuracy, recall, precision, and F-measure. Table 2 specifies the performance of classifiers, and Figure 7 illustrates the accuracy comparison between various classifiers.

Accuracy: Accuracy refers to the proportion of correctly classified instances among all instances. It is calculated by dividing the number of correct predictions by the total number of predictions. In the context of CHD prediction and diagnosis, accuracy measures how well the classifier can correctly identify patients with and without CHD.

$$\text{Accuracy} = (TP + TN) / (TP + TN + FP + FN)$$

Recall: Recall, also known as sensitivity or true positive rate, refers to the proportion of true positive instances among all positive instances. It is calculated by dividing the number of true positive predictions by the total number of positive instances. In the context of CHD prediction and diagnosis, recall measures how well the classifier can correctly identify patients with CHD, i.e., the percentage of actual CHD cases correctly identified by the classifier.

$$\text{Recall} = TP / (TP + FN)$$

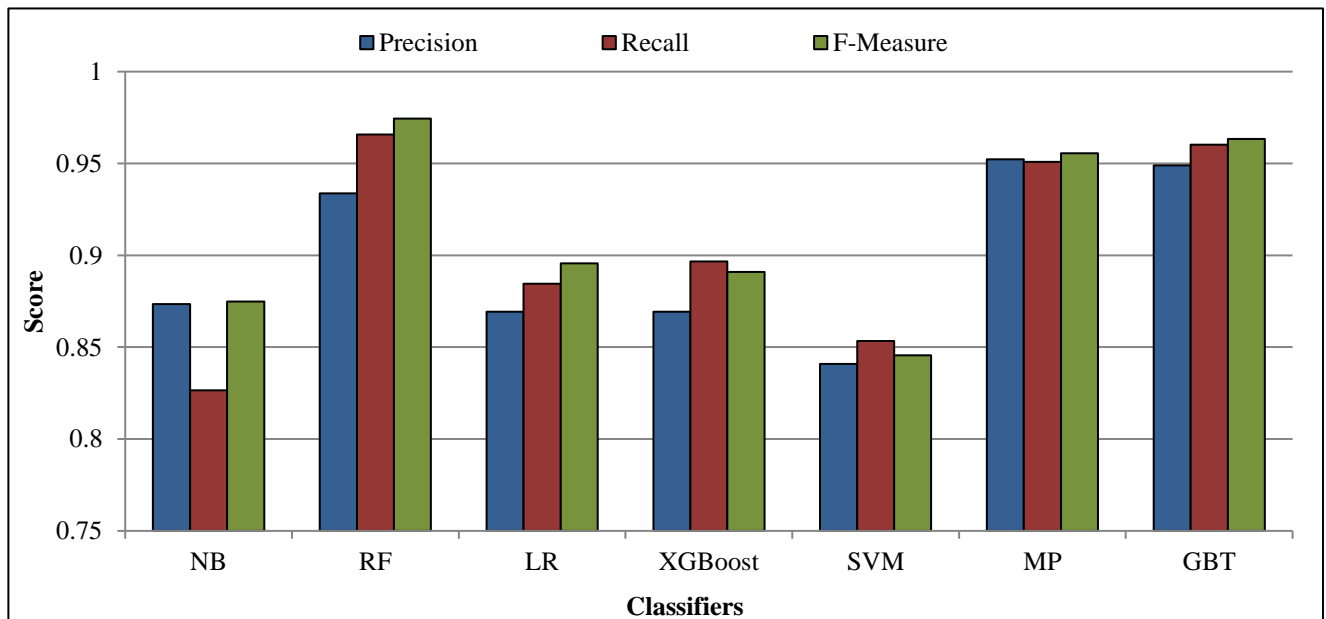

**Fig. 8 Precision, recall, and F measure comparison between various classifiers**

Precision: Precision refers to the proportion of true positive instances among all predicted positive instances. It is calculated by dividing the number of true positive predictions by the total number of predicted positive instances. In the context of CHD prediction and diagnosis, precision measures how well the classifier can correctly identify patients with CHD, i.e., the percentage of predicted cases of CHD that are actually true positive cases.

$$\text{Precision} = TP / (TP + FP)$$

F-measure: F-measure, also known as the F1 score, is the harmonic mean of precision and recall. It is calculated as the weighted average of precision and recall, where the weights are equal, and the formula is:

$$\text{F-measure} = 2 \times (\text{precision} \times \text{recall}) / (\text{precision} + \text{recall})$$

In the context of CHD prediction and diagnosis, F-measure combines the precision and recall metrics to give an overall performance score that considers both the true positive rate and the false positive rate. In addition to these metrics, other evaluation metrics can be used, such as specificity, negative predictive value, and area under the receiver operating characteristic (ROC) curve.

Specificity measures the proportion of true negative instances among all negative instances. It is calculated by dividing the number of true negative predictions by the total number of negative instances. The negative predictive value measures the proportion of true negative instances among all predicted negative instances. It is calculated by dividing the number of true negative predictions by the total number of predicted negative instances. The area under the ROC curve measures how well the classifier can distinguish between positive and negative instances and ranges from 0.5 (random guessing) to 1 (perfect classification).

When evaluating the performance of machine learning classifiers for CHD prediction and diagnosis, it is important to consider the trade-offs between different metrics. For example, a classifier with high recall but low precision may identify many patients as having CHD, including some false positives, while a classifier with high precision but low recall may identify only a small number of patients with CHD, missing some true positive cases. The choice of evaluation metric depends on the specific goals and requirements of the application, and different metrics may be appropriate for different stages of the diagnostic process.

The performance evaluation of various machine learning (ML) classifiers for predicting and diagnosing Coronary Heart Disease (CHD) using multiple metrics, including recall, precision, and F-measure. These metrics provide additional insights into the performance of a classifier beyond accuracy and are commonly used in classification tasks, especially in cases where the classes are imbalanced or where different types of errors may have different costs or consequences.

The evaluation results are presented in Figure 8, which presumably shows a comparison of the performance of various ML classifiers for predicting and diagnosing CHD using recall, precision, and F-measure. The passage notes that three specific classifiers, namely Random Forest, Gradient Boosting Tree, and Multilayer Perception, performed well in the evaluation, achieving scores of more than 95 in all three metrics. It is important to note that the performance of a classifier can vary depending on the dataset's characteristics, the experimental setup, and the specific evaluation metrics used. Therefore, it is important to select and report multiple metrics carefully and to validate the results using cross-validation or other techniques to ensure the robustness and reliability of the evaluation.

## 5. Conclusion

In conclusion, we present a comprehensive framework for foreseeing the occurrence of heart disease using ML classifiers. The proposed framework comprises various phases, including data analysis, feature selection, ReliefF, 3D CNN-based segmentation, feature extraction by means of transfer learning, feature fusion in addition to classification, and Adagrad optimization.

The use of machine learning classifiers for predicting and diagnosing Coronary Heart Disease (CHD) shows great promise in the field of medical diagnostics. This approach offers a high level of accuracy and can help medical professionals make more informed decisions when it comes to diagnosing and treating CHD. Through this article, we have explored various machine-learning techniques used to predict CHD, including Decision Trees, Random Forests, SVM, and Neural Networks. The overall findings have shown that several ML classifiers may acquire excellent precision in forecasting CHD, despite the fact that every one of these methods has certain advantages and disadvantages.

The outcomes demonstrate that the suggested strategy exceeds existing techniques and achieves excellent accuracy. The use of ReliefF for feature selection helps reduce the number of features and improve the classifier's performance. The 3D CNN-based segmentation helps to identify the regions of interest in the image; feature extraction also improves the classifier's accuracy. The feature fusion technique helps to combine the features extracted from different sources, and Adagrad optimization further improves the classifier's performance. The proposed framework can be used as a valuable tool for early diagnosis and treatment of coronary heart disease. Future studies can focus on improving the accuracy along with the efficiency of the proposed framework and extending it to other medical applications.

# References

[1]   R. Alizadehsani et al., "Coronary Artery Disease Detection using Deep Learning Methods," *Procedia Computer Science*, vol. 98, pp. 360-365, 2016.

[2]   Zachi I. Attia et al., "Prospective Validation of a Deep Learning Electrocardiogram Algorithm for the Detection of Left Ventricular Systolic Dysfunction," *Journal of Cardiovascular Electrophysiology,* vol. 30, no. 5, pp. 668-674, 2019. [CrossRef] [Google Scholar] [Publisher Link]

[3]   K. Sudharson et al., "Hybrid Deep Learning Neural System for Brain Tumor Detection," *2022 2nd International Conference on Intelligent Technologies (CONIT)*, pp. 1-6, 2022. [CrossRef] [Google Scholar] [Publisher Link]

[4]   L. Srinivasan et al., "IoT-Based Solution for Paraplegic Sufferer to Send Signals to Physician via Internet," *SSRG International Journal of Electrical and Electronics Engineering*, vol. 10, no. 1, pp. 41-52, 2023. [CrossRef] [Publisher Link]

[5]   Kwon JM, Lee YJ, and JS. Kim, "Machine Learning-based Prediction of Myocardial Infarction Using Big Data: A Nationwide Cohort Study," *BMC Cardiovascular Disorders,* vol. 20, no. 1, 2020.

[6]   X. Xie et al., "Automatic Detection of Coronary Artery Disease using a Deep Convolutional Neural Network with Noise Reduction and Bayesian Optimization," *Medical Image Analysis*, vol. 60, p. 101610, 2020.

[7]   P.R. Gajjala et al., "An Exploratory Analysis of Risk Factors for Coronary Heart Disease Using Logistic Regression Modeling," *Clinical Medicine & Research*, vol. 9, no. 2, pp. 107– 113, 2017.

[8]   Lu W, and K. Reshma, "Application of Logistic Regression Analysis in Predicting Coronary Heart Disease," *Journal of Taibah University Medical Science*, vol. 15, no. 4, pp. 289-295, 2020.

[9]   X. Chen et al., "Decision Trees for Predicting Coronary Heart Disease Based on Genetic and Clinical Data," *BMC Medical Genetics*, vol. 19, no. 1, pp. 1-13, 2018.

[10]  J. Wu et al., "Prediction of Coronary Heart Disease Risk Based on SVM," *International Journal of Clinical and Experimental Medicine*, vol. 13, no. 10, pp. 8038-8045, 2020.

[11]  N. Garg, D. Gupta, and A. Yadav, "Convolutional Neural Network for Detection of Coronary Artery Disease using Myocardial Perfusion Scintigraphy," *Computers in Biology and Medicine*, vol. 98, pp. 8-15, 2018.

[12]  S. Pujari, P. Rao, and S. Allamsetty, "Deep Neural Networks for Predicting the Risk of Heart Diseases," *International Journal of Advanced Computer Science Applications*, vol. 12, no. 2, pp. 361-368, 2021.

[13]  J. Huang et al., "An Ensemble of Logistic Regression Model to Predict Coronary Artery Disease for a Chinese Population," *Computational and Mathematical Methods in Medicine,* p. 2381843, 2019.

[14]  A. Hernáez et al., "Meta-analysis of Randomized Controlled Trials: Efficacy and Safety of Vitamin D Supplementation in Relation to HbA1c Concentrations in Prediabetes and Diabetes," *Nutrition, Metabolism and Cardiovascular Diseases,* vol. 31, no. 2, pp. 325-335, 2021.

[15]  X. Xu et al., "Diagnostic Accuracy of Serum Matrix Metalloproteinase-9 for Hepatocellular Carcinoma: A Systematic Review and Meta-Analysis," *European Review for Medical and Pharmacological Sciences,* vol. 24, no. 18, pp. 9662-9672, 2020.

[16]  Pia R. Kamstrup, and Borge G. Nordestgaard, "Elevated Lipoprotein(a) Levels, LPA Risk Genotypes, and Increased  Risk of Heart Failure in the General Population," *Journal of the American College of Cardiology: Heart Failure*, vol. 4, no. 1, pp. 78-87, 2016. [CrossRef] [Google Scholar] [Publisher Link]

[17]  B.B. Walker et al., "Spatial Association Between County-level Breast Cancer Mortality Rates and Timber Harvests, Pesticide Applications and Hazardous Waste Sites in California," *Spatial and Spatio-temporal Epidemiology*, vol. 27, pp. 67-77, 2018.

[18]  Joann E. Manson et al., "Vitamin D Supplements and Prevention of Cancer and Cardiovascular Disease," *The New England Journal of Medicine*, vol. 380, no. 1, pp. 33-44, 2019. [CrossRef] [Google Scholar] [Publisher Link]

[19]  Amit V. Khera et al., "Genetic Risk, Adherence to a Healthy Lifestyle, and Coronary Disease," *The New England Journal of Medicine*, vol. 375, pp. 2349-2358, 2016. [CrossRef] [Google Scholar] [Publisher Link]

[20]  V.E. Hjortdal et al., "Risk of Infective Endocarditis in Patients with Bicuspid Aortic Valve: A Population-Based Cohort Study," *The American Journal of Medicine*, vol. 130, no. 2, 2017.

[21]  S.A.H. Rizvi et al., "A Systematic Review of the Efficacy and Safety of Statins in the Pediatric Population," *American Journal of Cardiovascular Drugs*, vol. 20, no. 5, pp. 399-406, 2020.

[22]  Y. Guo et al., "Effect of Omega-3 Carboxylic Acids on Lipoprotein-associated Phospholipase A2 and Cardiovascular Events in Patients with Elevated Triglyceride Levels: A Meta-analysis of Randomized Controlled Trials," *Clinical Nutrition*, vol. 38, no. 4, pp. 1634-1641, 2019.

[23]  Y. Liu et al., "Relationship between Serum Vitamin D Level and Breast Cancer: A Meta-analysis of Observational Studies," *The Journal of Clinical Endocrinology & Metabolism*, vol. 103, no. 7, pp. 2970-2979, 2018.

[24] Williams Bryan et al., "2018 ESC/ESH Guidelines for the Management of Arterial Hypertension: The Task Force for the Management of Arterial Hypertension of the European Society of Cardiology and the European Society of Hypertension: The Task Force for the Management of Arterial Hypertension of the European Society of Cardiology and the European Society of Hypertension.," *Journal of Hypertension*, vol. 36, no. 10, pp. 1953-2041. 2018. [CrossRef] [Publisher Link]

[25] P. Bharadwaj et al., "Adherence to Mediterranean diet and risk of Stroke: A Systematic Review and a Meta-Analysis of Observational Studies," *Clinical Nutrition ESPEN*, vol. 15, pp. 47-57, 2016.

[26] D. Selvaraj et al., "Outsourced Analysis of Encrypted Graphs in the Cloud with Privacy Protection," *SSRG International Journal of Electrical and Electronics Engineering*, vol. 10, no. 1, pp. 53-62, 2023. [CrossRef] [Publisher Link]

[27] P Joe Prathap et al., "Mining Privacy-Preserving Association Rules based on Parallel Processing in Cloud Computing," *International Journal of Engineering Trends and Technology*, vol. 70, no. 3, pp. 284-294, 2022. [CrossRef] [Publisher Link]

[28] M Balachandran et al., "Predictive Analytics in Coronary Artery Disease - A Systematic Review," *Computer Methods and Programs in Biomedicine,* vol. 208, pp. 106237, 2021.

[29] S.M. Udhaya Sankar et al., "Safe Routing Approach by Identifying and Subsequently Eliminating the Attacks in MANET," *International Journal of Engineering Trends and Technology*, vol. 70, no. 11, pp. 219-231, 2022. [CrossRef] [Publisher Link]

[30] SC Chien et al., "Prediction of Coronary Artery Disease Using Machine Learning: An Experience of Integrating Electronic Medical Claims and Clinical Data," *PLoS One*, vol. 16, no. 2, 2021.

[31] Jena Catherine Bel D et al., "Trustworthy Cloud Storage Data Protection based on Blockchain Technology," *International Conference on Edge Computing and Applications (ICECAA),* pp. 538-543, 2022, [CrossRef] [Google Scholar] [Publisher Link]

[32] Shakti Chourasiya, and Suvrat Jain, "A Study Review on Supervised Machine Learning Algorithms," *SSRG International Journal of Computer Science and Engineering*, vol. 6, no. 8, pp. 16-20, 2019. Crossref, [CrossRef] [Publisher Link]

[33] G. Gomathy, "Automatic Waste Management based on IoT using a Wireless Sensor Network," *International Conference on Edge Computing and Applications (ICECAA),* pp. 629-634, 2022. [CrossRef] [Google Scholar] [Publisher Link]

[34] He B et al., "A Machine Learning Approach for the Diagnosis of Coronary Artery Disease Based on Clinical Data," *Journal of Biomedical Informatics,* vol. 81, pp. 202-207, 2018.

[35] G. R. Meghana, Suresh Kumar Rudrahithlu, and K. C. Shilpa, "Detection of Brain Cancer using Machine Learning Techniques a Review," *SSRG International Journal of Computer Science and Engineering ,* vol. 9, no. 9, pp. 12-18, 2022. [CrossRef] [Publisher Link]

[36] D. Dhinakaran, and P. M. Joe Prathap, "Preserving Data Confidentiality in Association Rule Mining Using Data Share Allocator Algorithm," *Intelligent Automation & Soft Computing*, vol. 33, no.3, pp. 1877–1892, 2022. [CrossRef] [Google Scholar] [Publisher Link]

[37] N Khalifa et al., "A Hybrid Deep Learning Approach for Coronary Artery Disease Diagnosis," *Journal of Biomedical Informatics,* vol. 109, pp. 103547, 2020.

[38] D. Dhinakaran et al., "Secure Android Location Tracking Application with Privacy Enhanced Technique," *Fifth International Conference on Computational Intelligence and Communication Technologies (CCICT),* pp. 223-229, 2022. [CrossRef] [Google Scholar] [Publisher Link]

[39] S. Farjana Farvin, and S. Krishna Mohan, "A Comparative Study on Lung Cancer Detection using Deep Learning Algorithms," *SSRG International Journal of Computer Science and Engineering ,* vol. 9, no. 5, pp. 1-4, 2022. [CrossRef] [Publisher Link]

[40] J. Aruna Jasmine et al., "A Traceability set up using Digitalization of Data and Accessibility," *International Conference on Intelligent Sustainable Systems (ICISS)*, pp. 907-910, 2021. [CrossRef] [Google Scholar] [Publisher Link]

[41] R. Tamilaruvi et al., "Brain Tumor Detection in MRI Images using Convolutional Neural Network Technique," *SSRG International Journal of Electrical and Electronics Engineering,* vol. 9, no. 12, pp. 198-208, 2022. [CrossRef] [Publisher Link]

[42] D. Dhinakaran et al., "Recommendation System for Research Studies Based on GCR," *International Mobile and Embedded Technology Conference (MECON),* pp. 61-65, 2022. [CrossRef] [Google Scholar] [Publisher Link]

[43] MY Rashwan et al., "Coronary Artery Disease Diagnosis Using Machine Learning Algorithms: A Review," *Journal of Medical Systems,* vol. 41, no. 9, pp. 139, 2017.

[44] D. Dhinakaran, and P. M. Joe Prathap, "Ensuring Privacy of Data and Mined Results of Data Possessor in Collaborative ARM," *Pervasive Computing and Social Networking, Lecture Notes in Networks and Systems*, vol. 317, pp. 431–444, 2022. [CrossRef] [Google Scholar] [Publisher Link]

[45] S. Rajeswari et al., "Aspect Based Polarity Extraction in Tamil Tweets using Tree-Based Recursive Partitioning Techniques," *International Journal of Engineering Trends and Technology*, vol. 70, no. 12, pp. 421-430, 2022. [CrossRef] [Publisher Link]

[46] L Yan et al., "Early Detection and Classification of Coronary Artery Disease using Machine Learning Techniques," *Journal of Healthcare Engineering,* vol. 2018, pp. 2581023, 2018.

[47]  Dhinakaran D, and Joe Prathap P. M, "Protection of Data Privacy From Vulnerability using Two-Fish Technique with Apriori Algorithm in Data Mining," *The Journal of Supercomputing*, vol. 78, no. 16, pp. 17559–17593, 2022. [CrossRef] [Google Scholar] [Publisher Link]

[48]  D Dhinakaran et al., "Assistive System for the Blind with Voice Output Based on Optical Character Recognition*," International Conference on Innovative Computing and Communications,  Lecture Notes in Networks and Systems,* vol. 492, pp. 1-8, 2022. [CrossRef] [Google Scholar] [Publisher Link]

[49]  H Ghafoori et al., "Prediction of Coronary Artery Disease using a Machine Learning Approach Based on Feature Selection of Echocardiographic Variables," *Journal of Medical Systems*, vol. 43, no. 6, pp. 139, 2019.

[50]  W He et al., "Prediction of Coronary Artery Disease using a Combination of Convolutional Neural Networks and Long Short-Term Memory with Comprehensive Features from Electronic Health Records," *Journal of Medical Systems*, vol. 45, no. 2, pp. 19, 2021.